\documentclass[10pt,twocolumn,letterpaper]{article}
\pdfoutput=1
\usepackage{cvpr}
\usepackage{times}
\usepackage{epsfig}
\usepackage{graphicx}
\usepackage{amsmath}
\usepackage{amssymb}
\usepackage{bbm}
\usepackage{comment}
\usepackage[ruled,lined]{algorithm2e}

\DeclareMathOperator*{\argmax}{argmax}

\newcommand{\lu}[1]{{\color{blue}{(Lu\@: #1)}}}

\newcommand{\llc}[1]{{\color{red}{(llc\@: #1)}}}

\usepackage[pagebackref=true,breaklinks=true,letterpaper=true,colorlinks,bookmarks=false]{hyperref}

\cvprfinalcopy 

\def\cvprPaperID{1240} 

\ifcvprfinal\fi
\begin{document}

\setlength{\abovedisplayskip}{2pt}
\setlength{\belowdisplayskip}{2pt}
\setlength{\textfloatsep}{2pt}

\title{
Focal Visual-Text Attention for Visual Question Answering}

\author{
Junwei Liang\textsuperscript{1} \qquad
Lu Jiang\textsuperscript{2} \qquad
Liangliang Cao\textsuperscript{3} \qquad
Li-Jia Li\textsuperscript{2} \qquad
Alexander Hauptmann\textsuperscript{1}  \vspace{.3em}\\
\textsuperscript{1}Carnegie Mellon University \qquad\qquad \textsuperscript{2}Google Inc. \qquad\qquad \textsuperscript{3}HelloVera AI\\
{\tt\small \{junweil,alex\}@cs.cmu.edu, \{lujiang,lijiali\}@google.com, liangliang.cao@gmail.com} 
\vspace{-.5em}\\
}

\maketitle

\begin{abstract}
\vspace{-2mm}
Recent insights on language and vision with neural networks have been successfully applied to simple single-image visual question answering. However, to tackle real-life question answering problems on multimedia collections such as personal photos, we have to look at whole collections with sequences of photos or videos. When answering questions from a large collection, a natural problem is to identify snippets to support the answer. In this paper, we describe a novel neural network called Focal Visual-Text Attention network (FVTA) for collective reasoning in visual question answering, where both visual and text sequence information such as images and text metadata are presented. FVTA introduces an end-to-end approach that makes use of a hierarchical process to dynamically determine what media and what time to focus on in the sequential data to answer the question. FVTA can not only answer the questions well but also provides the justifications which the system results are based upon to get the answers. FVTA achieves state-of-the-art performance on the MemexQA dataset and competitive results on the MovieQA dataset.
\end{abstract}

\vspace{-4mm}
\section{Introduction}


Language and vision have emerged as a popular research area in computer vision. Visual question answering (VQA)~\cite{antol2015vqa} is a successful direction utilizing both computer vision and natural language processing techniques to solve an interesting problem: given a pair of image and a question (in natural language), the goal is to learn an inference model that can the answer questions according to cues discovered from the image. A variety of methods have been proposed to address the challenges from different aspects~\cite{fukui2016multimodal,yang2016stacked,malinowski2015ask,nips15_hermann,shih2016look,ben2017mutan,nam2016dual,lu2016hierarchical}, with remarkable progress on answering about a single image.

\begin{figure}[t!]
	\centering
		\includegraphics[width=0.47\textwidth]{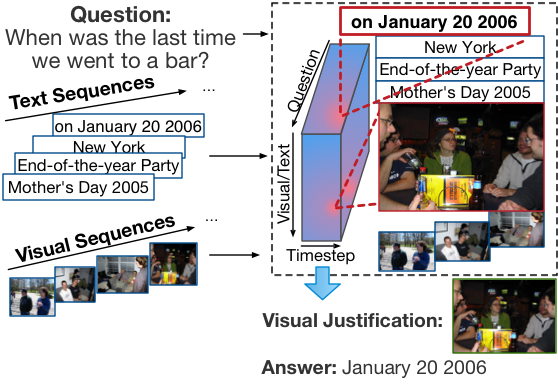}
	\caption{Focal Visual-Text Attention (FVTA) Mechanism.  Given the visual-text sequences input and the question, our temporal visual-text attention tensor captures the temporal constraint in the question and emphasizes the most recent image with "bar" scene visible. Then FVTA selects the appropriate attention region (the ``date") and finds the correct answer.}
	\label{FVT}
\end{figure}

Extending from VQA  on a single image, 
this paper considers the following 
problem: Suppose a user's photos and videos are organized in a sequence ordered by their creation time. Some photos or videos may be associated with meta labels or annotations such as time, GPS, captions, comments, and meaningful title. We are interested in training a model to answer questions about these images and texts, \eg ``when was the last time I went to a bar?" or ``what did my son do after his 2017 Halloween dinner party?"

There are two challenges to solve the above problem. 
First, the input is provided in an unstructured form. The question is associated with multiple sequences, in the form of videos or images. Such sequences are temporally ordered, and each sequence contains multiple time steps. At each time there are visual data, text annotations and other metadata. In this paper, we call the format 
\textbf{\textit{visual-text sequence}} data. Note that not all the photos and videos are annotated, which requires a robust method to leverage inconsistently available multimodal data. 




The second challenge requires interpretable justifications in addition to direct answer based on sequence data. To help users with a lot of photos and videos, 
a natural requirement is to identify the supporting evidence for the answer. An example question
as shown in Fig.~\ref{FVT}, is  ``when was the last time I went to a bar?" From the users' viewpoint, a good QA system should not only give a definite answer (\eg, January 20, 2016), but also
ground evidential images or text snippets in the input sequence to justify the reasoning process. Given imperfect VQA models, humans often want to verify the answer. The inspection process may be trivial for a single image but can take a significant amount of time to examine every image and the complete text words. 

To address these two challenges, we propose a focal visual-text attention (FVTA) model for sequential data~\footnote{Code and models are released at \url{https://memexqa.cs.cmu.edu/fvta.html}}. 
Our model is motivated by the reasoning process of humans. In order to answer a question, a human would first quickly skim the input and then focus on a few, small temporal regions in the visual-text sequences to derive an answer. In fact, statistics suggest that, on average, humans only need 1.5 images to answer a question after the skimming~\cite{jiang2017memexqa}. Inspired by this process, FVTA first learns to localize relevant information within \emph{a few, small, temporally consecutive regions} over the input sequences, and learns to infer an answer based on the cross-modal statistics pooled from these regions. FVTA proposes a novel kernel to compute the attention tensor that jointly models the latent information in three sources: 1) answer-signaling words in the question, 2) temporal correlation within a sequence, and 3) cross-modal interaction between the text and image. FVTA attention allows for collective reasoning by the attention kernel learned over a few, small, consecutive sub-sequences of text and image. It can also produce a list of evidential images/texts to justify the reasoning. As shown in Fig.~\ref{FVT}, the highlighted cubes are regions of high activations in the proposed FVTA. To summarize, the contribution of this paper is threefold:

\begin{figure*}[!t]
	\centering
		\includegraphics[width=0.97\textwidth]{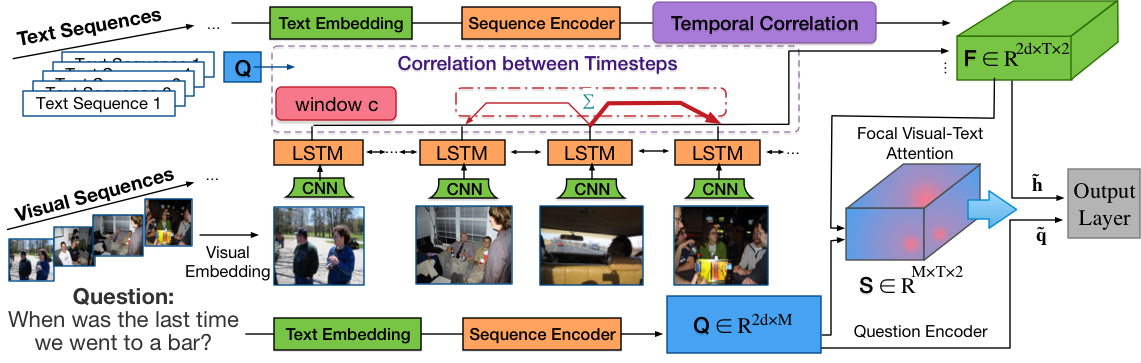}
	\caption{\label{fig:model_overview}An overview of Focal Visual-Text Attention (FVTA) model. For visual-text embedding, we use a pre-trained convolutional neural network to embed the photos and pre-trained word vectors to embed the words. We use a bi-directional LSTM as the sequence encoder. All hidden states from the question and the context are used to calculate the FVTA tensor. 
	Based on the FVTA attention, both question and the context are summarized into single vectors for the output layer to produce final answer. The output layer is used for multiple choice question classification. The text embedding of the answer choice is also used as the input. This input is not shown in the figure.
	} 
	\label{FVT-p2}
\vspace{-4mm}
\end{figure*}

\begin{itemize}
\vspace{-2mm}
    \item We propose a novel attention kernel for VQA on visual-text data. Experiments show that it outperforms existing attention methods. 
\vspace{-2mm}
    \item The proposed attention tensor can be used to localize evidential image and text snippets to explain the reasoning process. We quantitatively verify that the evidence produced by our method are more correlated to that of human annotators.
\vspace{-2mm}
    \item Our method achieves the state-of-the-art results on two VQA benchmarks.
\end{itemize}

\section{Related Work}


\textbf{Visual Question Answering.} 
Image-based visual question answering has received a large amount of interest in the computer vision community. 
A lot of efforts have been conducted on single image QA datasets \cite{antol2015vqa,krishna2017visual,zhu2016visual7w,noh2016image,xu2016ask,andreas2016learning}, where a common practice is to train a classifier by combining both question feature and visual features.
A recent direction is on the
question answering based on videos, which is more relevant to this work.
A number of research studies have been carried on MovieQA \cite{tapaswi2016movieqa,kim2017deepstory,na2017read}, with movie clips, scripts, and descriptions.
Because it is expensive to annotate the video-based QA datasets, 
some research studies generate QA datasets by harvesting online videos and descriptions
\cite{zhu2015uncovering,zeng2017leveraging}, while a recent study \cite{jang2017tgif} considers question answering using animated GIFs. 
This work differs from the existing video-based QA in two aspects: (1) video-based QA is to answer questions based on a single video, while our work can handle general visual-text sequences, where one user may have more than one video or albums of photos. (2) most existing video-based QA methods map one video sequence with text into a context feature vector, while our work explores a more fine-grained model by modeling the correlation between query and sequence data at every time step. To this end, we experiment on the MemexQA dataset \cite{jiang2017memexqa}. The sequential data in MemexQA involves multiple modalities, including titles, timestamps, GPS and visual content, render it an ideal test bed for QA research over visual-text sequence data. Unlike the model in \cite{jiang2017memexqa}, our method also uses the text embedding of the answer choices as the input to answer a question.





\textbf{Attention Mechanism.}
This work can be viewed as a novel attention model for multiple variable-length sequential inputs, to take into account not only the visual-text information but also the temporal dependency. 
Our work extends the previous studies of using attention model for Image QA \cite{shih2016look,das2017human,xu2016ask,lu2016hierarchical,yang2016stacked,nam2016dual,fukui2016multimodal,ben2017mutan}.
A key difference between our method and classical attention model lies in the fact we are modeling the correlation at every time step, across multiple sequences.  Existing attention mechanisms for VQA mainly focus on attention within spatial regions of an image~\cite{zhu2016visual7w} or within a single sequence~\cite{jang2017tgif}, and hence, may not fully exploit the multiple sequences and multiple time steps nature. As Fig.~\ref{FVT_comp} shows, our attention is applied to a three-dimensional tensor, while the classic soft attention model is applied to a vector or matrix.

\section{Approach}

\subsection{Problem Formulation}
We start the discussion by formally defining the problem. Let $Q = q_1, \cdots, q_M$ represent a question of $M$ words $Q \in \mathbb{Z}^M$, where each word is an integer index in the vocabulary. Define a context visual-text sequence of $T$ examples $\mathbf{X} = \mathbf{x}_1, \cdots, \mathbf{x}_T$, where for each example, $\mathbf{x}_t^{img}$ represents an image. $\mathbf{x}_t^{txt}$ is its corresponding text sentence, where its $i$-th word is indexed by $\mathbf{x}_{ti}^{txt}$. Following~\cite{antol2015vqa,zhu2016visual7w}, the answer to a question is an integer $y \in [1,L]$ over the answer vocabulary of size $L$. Given a collection of $n$ questions and their context sequences, we are interested in learning a model maximizing the following likelihood:
\begin{equation}
\argmax_{\Theta} \sum_{i=1}^n \log P(y_i | Q_i,\mathbf{X}_i;\Theta)
\end{equation}
where $\Theta$ represents the model parameters. Given the visual-text sequence input $\mathbf{X}^{img},\mathbf{X}^{txt}$, we obtain a good joint representation by attention model. With FVTA attention, the model takes into account of the sequential dependency in image or text sequence, respectively, and cross-modal visual-text correlations. Meanwhile, the computed attention weights over input sequences can be utilized to derive meaningful justifications.


\subsection{Network Architecture}

This subsection discusses our overall neural network architecture.
As shown in Fig.~\ref{fig:model_overview}, the proposed network consists of the following layers. 


\noindent\textbf{Visual-Text Embedding} Every image or video frame is encoded 
with a pre-trained Convolutional Neural Network.
Both word-level and character level embedding~\cite{kim2014convolutional} are used to represent the word in text and question. 



\noindent\textbf{Sequence Encoder}
We use separate LSTM networks to encode visual and text sequences, respectively, to capture the temporal dependency within each individual sequence. The inputs to the LSTM units are image/text embedding produced by the previous layer. Let $d$ denote the size of the hidden state of the LSTM unit; the question $Q$ is represented as a matrix $\mathbf{Q}$ of concatenated bi-directional LSTM outputs at each step, \ie, $\mathbf{Q} \in \mathbb{R}^{2d \times M}$, where $M$ is the maximum length of the question. Likewise, The sequentially encoded text and images are represented by $\mathbf{H} \in \mathbb{R}^{2d \times T \times 2}$, where $T$ is the maximum length of the sequence.

\noindent\textbf{Focal Visual-Text Attention}
The FVTA is a novel layer to implement the proposed attention mechanism. It represents a network layer that models the correlations between questions and multi-dimensional context and produces the summarized input to the final output layer, \textit{i.e.}, $\mathbf{\tilde{h}} \in \mathbb{R}^{2d}$ and $\mathbf{\tilde{q}} \in \mathbb{R}^{2d}$. We will discuss FVTA in the next section.


\noindent\textbf{Output Layer}
After summarizing the input using the FVTA attention, we use a feed-forward layer to obtain the answer candidate. For multiple-choices questions, the task is to select one answer from a few candidate choices given the context and the question. Let $k$ denote the number of candidate answers, we utilize the bi-directional LSTM to encode each of the answer choice and use the last hidden state as the representation for answers $\mathbf{E} \in \mathbb{R}^{k \times 2d}$. We tile the context representation $\mathbf{\tilde{h}}$ and attended question representation, $k$ times into $\mathbf{\tilde{H}} \in \mathbb{R}^{k \times 2d}$ and $\mathbf{\tilde{Q}} \in \mathbb{R}^{k \times 2d}$ to compute  
the classification probability of $k$ choices. In practice we find the following simple equation works better than fully connected layer or straightforward concatenation:

\begin{equation} \label{eq:output}
\mathbf{p} = softmax(\mathbf{w_p^T}[\mathbf{\tilde{Q}} ; \mathbf{\tilde{H}} ; \mathbf{E} ; \mathbf{\tilde{Q}} \odot \mathbf{E}; \mathbf{\tilde{H}} \odot \mathbf{E}])
\vspace{1mm}
\end{equation}
where the operator $[ \cdot ; \cdot ]$ represents the concatenation of two matrices along the last dimension. $\odot$ is the element-wise multiplication, $\mathbf{w_p}$ is the weight vector to learn and $\mathbf{p}$ is a vector of classification probability. 
After obtaining the answer probability, the model can be trained end-to-end using cross-entropy loss function.

\label{sec:approach}

\section{Focal Visual-Text Attention}

This section discusses the details of FVTA model as the key module in our VQA system. We first introduce similarity metric between visual and text features, then discuss constructing the attention tensor that captures both intra-sequence dependency and inter-sequence interaction.


\subsection{Similarity between visual and text features}

To compute the similarity across different modalities, \ie visual and text, we first encode every modality by the LSTM networks with the same size of hidden states. Then we measure the differences between these hidden state variables. Following the study in text sequence matching~\cite{wang2016compare}, we aggregate both the cosine similarity and Euclidean distance to compare the features. Moreover, we choose to keep the vector information instead of summing up after the operation. The vector representation can be used as the input of a learning model, whose inner product represents the similarity between these features. 
More specifically, we use the following equation to compute the similarity representation between two hidden state vectors $\mathbf{v}_1$ and $\mathbf{v}_2$. The result is a vector of twice the hidden size:
\begin{equation}\label{eq-aggregate-similarity}
\mathbf{s}(\mathbf{v}_1,\mathbf{v}_2) =[(\mathbf{v_1}\odot\mathbf{v}_2) ; (\mathbf{v}_1-\mathbf{v}_2)\odot
(\mathbf{v}_1-\mathbf{v}_2)].
\end{equation}

\subsection{Intra-sequence temporal dependency}

Our visual-text attention layer is designed to let the model select related visual-text region or timestep based on each word of the question. Such fine-grained attention is in general nontrivial to learn. Meanwhile, most answers for visual-text sequence inputs may be constrained and restricted in a short temporal period.
We learn such localized representation, called focal context representation, to emphasize relevant context states based on the question.

First, we introduce a temporal \textit{correlation matrix}, $\mathbf{C} \in \mathbb{R}^{T \times T}$, a symmetric matrix where each entry $c_{ij}$ measures the correlation between context's the $i$-th step and the $j$-th step for a question. Let $\mathbf{h}_{i} = \mathbf{H}_{:i:} \in \mathbb{R}^{2d \times 2}$ denote the visual/text representation for the $i$-th timestep in $\mathbf{H}$. For notation convenience, $:$ is a slicing operator to extracts all elements from a dimension. 
For example, $\mathbf{h}_{i1} = \mathbf{H}_{:i1}$ represents the vector representation of the $i$-th timestep of the visual sequence. 
Here we denote the last index $1$ for visual and $2$ for textual modality.
Each entry $\mathbf{C}_{ij}$ ($\forall i, j \in [1,T]$) is then calculated by:
\begin{equation} \label{eq:time_w}
\begin{split}
\mathbf{C}_{ij} = \tanh\sum_{k=1}^2 \mathbf{w}_c^\top(\mathbf{w}^\top_h \mathbf{s}(\mathbf{h}_{ik}, \mathbf{h}_{jk})
+  \mathbf{Q}_{:M})
\end{split}
\end{equation}
where $\mathbf{w}_c \in \mathbb{R}^{2d \times 1}$ and $\mathbf{w}_h \in \mathbb{R}^{4d \times 2d}$
are parameters to learn. 
The temporal correlation matrix captures the temporal dependency of question, image and text sequence.


To allow the model to capture the context between timesteps based on the question, we introduce temporal focal pooling to connect neighboring time hidden states if they are related to the question.
For example, it can capture the relevance between the moment "dinner" and the moment later, "Went dancing", given the question "What did we do after the dinner on Ben's birthday?". 
Formally, given the time correlation matrix $\mathbf{C}$ and the context representation $\mathbf{H}$, we introduce a \textit{temporal focal pooling function} $g$ to obtain the focal representation $\mathbf{F} \in \mathbb{R}^{2d \times T \times 2}$. Each vector entry $\mathbf{F}_{:tk}$ ($\forall t \in [1,T], \forall k \in [1,2]$) in $\mathbf{F}$ is calculated by:
\begin{equation} \label{eq:time_g}
\mathbf{F}_{:tk} = g(\mathbf{H};\mathbf{C},t,k) \in \mathbb{R}^{2d},
\end{equation}
\begin{equation} \label{eq:time_g_h}
g(\mathbf{H};\mathbf{C},t,k) = \sum_{s=1}^T\mathbbm{1}[s \in [t-c,t+c]]\mathbf{C}_{st}\mathbf{h}_{sk},
\vspace{-1mm}
\end{equation}
where $\mathbf{F}_{:tk}$ is the focal context representation at $t$-th timestep for visual ($k=1$) or text ($k=2$). $\mathbbm{1}$ is the indicator function. $c$ stands for the size of the temporal window) that is end-to-end learned with other parameters. We constrain the model to focus on a few small temporal context windows of learnable window size $2c+1$.

\begin{figure}[!t]
	\centering
		\includegraphics[width=0.45\textwidth]{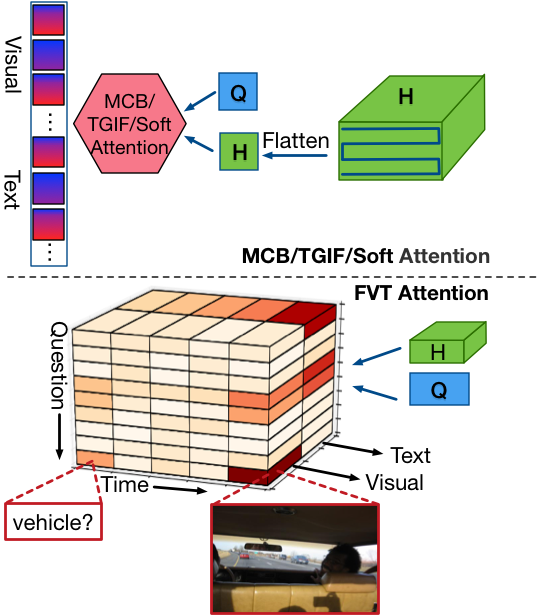}
	\caption{Comparison of our FVTA and classical VQA attention mechanism. FVTA considers both visual-text intra-sequence correlations and cross sequence interaction, and focuses on a few, small regions. In FVTA, the multi-modal feature representation in the sequence data is preserved without losing information.}
	\label{FVT_comp}
\end{figure}


\subsection{Cross Sequence Interaction}
In this section, we introduce the attention mechanism to capture the important correlation between visual and textual sequences. 
We apply attention over the focal context representation to summarize important information for answering the question.
We obtain the attention weights based on a correlation tensor $\mathbf{S}$ between each word of the question and each timestep of the visual-text sequences.
The attention at each timestep only considers the context and the question, and does not depend on the attention at previous timestep. The intuition of using such a memory-less attention mechanism is that it simplifies the attention and let the model focus on learning the correlation between context and question. Such mechanism has been proven useful in text question answering~\cite{seo2016bidirectional}.
We computes a \textit{kernel tensor}, $\mathbf{S} \in \mathbb{R}^{M \times T \times 2}$, between the input question and the focal context representation $\mathbf{F}$, where each entry in the kernel $s_{mtk}$ models the correlation between the $m$-th word in question and at $t$-th timestep over the modal $k$ (images or text words). Let $\mathbf{v}_{tk}$ denote the focal context representation $\mathbf{F}_{:tk}$ at $t$-th timestep for visual or text. Each entry $s_{mtk}$ in $\mathbf{S}$ is calculated by:
\begin{equation}    \label{eq:simi}
\begin{split}
s_{mtk} &= \kappa (\mathbf{F}_{:tk},\mathbf{Q}_{:m}) = \kappa(\mathbf{v}_{tk},\mathbf{q}) \\
&= \tanh(\mathbf{w^\top_s}\mathbf{s}(\mathbf{v}_{tk}, \mathbf{q})  + \mathbf{b}_s)
\end{split}
\end{equation}
where $\kappa$ is a function to compute the correlation between question and context, $\mathbf{w}_s \in \mathbb{R}^{4d \times 1}$ is the learned weights and $\mathbf{b}_s$ is the bias term. $\mathbf{s}$ is the mapping defined in \eqref{eq-aggregate-similarity}.
As explained for Eq.~\eqref{eq:time_w}, we use such similarity representations since they capture both the cosine similarity and Euclidean distance information.
We obtain the visual-text sequence attention matrix $\mathbf{A} \in \mathbb{R}^{T \times 2}$ by $\mathbf{A} = softmax(\max_{i=1}^M (\mathbf{S}_{i::}))$ and the visual-text attention vector  $\mathbf{B}  \in \mathbb{R}^{2}$ by $\mathbf{B} = softmax( \max_{i=1}^T\max_{j=1}^M (\mathbf{S}_{ji:}))$, where the softmax operation is applied to the first dimension.
The maximum function $\max_{i}$ is used reduce the first dimension of the high-dimensional tensor. Then the attended context vector is given by:
\begin{equation} \label{eq:attend_h}
\mathbf{\tilde{h}} = \sum_{k=1}^2\mathbf{B}_{k} \sum_{t=1}^T \mathbf{A}_{tk}\mathbf{F}_{:tk} \in \mathbb{R}^{2d}
\end{equation}
The visual-text attention is computed based on the correlation between question and the focal context attention, which aligns with our observation that questions often provide constrains of a limited time window for the answers.
Similarly, we compute the question attention  $\mathbf{D}  \in \mathbb{R}^{M}$ by $\mathbf{D} = softmax( \max_{i=1}^T\max_{j=1}^2 (\mathbf{S}_{:ij}))$  and the summarized question vector is given by:
\begin{equation} \label{eq:q_tilde}
\mathbf{\tilde{q}} = \sum_{m=1}^M\mathbf{D}_{m}\mathbf{Q}_{:m} \in \mathbb{R}^{2d}
\end{equation}
Algorithm~\ref{alg:overall} summarizes the steps to compute the proposed FVTA attention. To obtain a final context representation, we first summarize the focal context representation separately for visual sequence and text sequence, emphasizing the most important information using the intra-sequence attention. Then, we obtain the final representation by summing the sequence vector representation based on the inter-sequence importance. Fig.~\ref{FVT_comp} illustrates the difference between FVT attention tensor and one-dimensional soft attention vector. Both mechanisms compute the attention but FVTA considers both visual-text intra-sequence correlations and cross sequence interaction.

{
\begin{algorithm}[h]
\SetKwData{Left}{left}\SetKwData{This}{this}\SetKwData{Up}{up}
\SetKwInOut{Input}{input}\SetKwInOut{Output}{output}
\LinesNumbered
\Input{Input visual-text sequence $\mathbf{X}$, Question $Q$}
\Output{The FVTA vector $\mathbf{\tilde{h}}$}
\BlankLine
Encode $\mathbf{X}$ into $\mathbf{H}$ by the visual-text embedding and sequence encoder in Sec.~\ref{sec:approach}\;
Encode Q into $\mathbf{Q}$ by the question encoder\;
Compute $\mathbf{C}$ by Eq.~\eqref{eq:time_w} {\scriptsize\tcp{temporal correlation}} 
Compute $\mathbf{F}$ by Eq.~\eqref{eq:time_g}  {\scriptsize\tcp{intra-sequence dependency}} 
Compute $\mathbf{S}$ by Eq.~\eqref{eq:simi} {\scriptsize\tcp{cross-sequence interaction}}
Reduce $\mathbf{F}$ with $\mathbf{S}$ to the FVTA $\mathbf{\tilde{h}}$ by Eq.~\eqref{eq:attend_h}\;
\Return $\mathbf{\tilde{h}}$\;
\caption{\label{alg:overall} {\small Computation of Focal Visual-Text Attention.}}
\end{algorithm}
}


\label{sec:fvta}

\begin{table*}[]
\centering
\begin{tabular}{l||c|c|c|c|c|c}
\hline
Method            & how many & what & when & where  & who  & overall \\ 
&(11.8\%)  & (41.9\%)& (16.2\%)  &(17.2\%)&(12.9\%)&
\\\hline \hline
Logistic Regression             & 0.645 & 0.241 & 0.217 & 0.277 & 0.260 & 0.295 \\
Embedding + LSTM             & 0.771 & 0.564& 0.349& 0.314& 0.310& 0.478 \\
Embedding + LSTM + Concat            & 0.776 &	0.668& 	0.398& 	0.433	& 0.409	& 0.563 \\
\hline
DMN+  \cite{xiong2016dynamic}        & 0.792 & 0.616 & 0.346 & 0.248 & 0.224 & 0.480 \\
Multimodal Compact Bilinear Pooling \cite{fukui2016multimodal}& 0.773 & 0.618 & 0.250 & 0.229 & 0.248 & 0.462 \\

Bi-directional Attention Flow    \cite{seo2016bidirectional}    &0.790 & 0.689 & 0.356 & 0.567 & 0.468 & 0.598\\
Soft Attention          &  \textbf{0.795} & 0.697 & 0.346 & 0.604 & 0.582 & 0.621 \\
TGIF Temporal Attention   \cite{jang2017tgif}       & 0.761 & 0.700 & \textbf{0.522} & 0.582 & 0.477 & 0.630 \\
FVTA          &  0.761 & \textbf{0.714} & 0.476 & \textbf{0.676} & \textbf{0.668} & \textbf{0.669} \\
 \hline
\end{tabular}
\vspace{1mm}
\caption{Comparison of different methods on MemexQA by question type. The first three methods do not use the attention mechanism.}
\label{exp-mem}
\vspace{-4mm}
\end{table*}













\section{Experiments}\label{exp}


\subsection{MemexQA}
\noindent\textbf{Dataset} 
MemexQA~\cite{jiang2017memexqa} is a recently proposed visual-text question answering dataset. The dataset consists of 20,860 questions about 13,591 personal photos belonging to 101 real Flickr users. These personal photos capture a variety of key moments of their lives such as a trip to Japan, wedding ceremonies, family parties, etc. Each album and photo come with comprehensive visual-text information, including a timestamp, GPS, a photo title, an album title and description. The metadata is incomplete and GPS, the photo title, the album title and description may not present in every photo.

MemexQA provides 4 answer choices and only one correct answer for each question. The dataset also provides more than one ground truth grounding images for each question. There are five types of questions corresponding to the frequent search terms discovered in the Flickr search logs~\cite{jiang2017delving}.
The input visual-text sequence length varies for questions. Some questions are about images taken on a certain date \eg ``what did we do after 2006 Halloween party?''; others are about all images \eg ``what was the last time we drove to a bar?''.


\noindent\textbf{Baseline Methods} A large proportion of the existing solutions is to project image or videos into an embedding space, and train a classification model using these embeddings. We implement the following methods as baselines: \textbf{\textit{Logistic Regression}} predicts the answer with concatenated image, question and metadata features as reported in \cite{jiang2017memexqa}. \textbf{\textit{Embedding + LSTM}} utilizes word embeddings and character embeddings, along with the same visual embeddings used in FVTA. Embeddings are encoded by LSTM and averaged to get the final context representation. \textbf{\textit{Embedding + LSTM + Concat}} concatenates the last LSTM output from different modalities to produce the final output. On the other hand, we compare the proposed model to a rich collection of VQA attention models: \textbf{\textit{Classic Soft Attention}} uses classic one dimensional question-to-context attention to summarize context for question answering. A correlation matrix between each question word and context is used to compute the attention as in \cite{seo2016bidirectional,xu2016ask}. \textbf{\textit{DMN+}} is the improved dynamic memory networks \cite{xiong2016dynamic}, which is one of the representative architectures that achieve good performance on the VQA Task. We implement the DMN+ network with each sentence and each photo representation used in our proposed network as supporting facts input. 
\textbf{\textit{Multimodal Compact Bilinear Pooling}}\cite{fukui2016multimodal} is the state-of-the-art method on VQA~\cite{antol2015vqa} dataset. The spatial attention in the original model is directly used on the sequential images input. The hyperparameters including the output dimension of MCB and hidden size of LSTM are selected based on the validation results. 
\textbf{\textit{Bi-directional Attention Flow}} implements the single-modal attention flow model \cite{seo2016bidirectional} over all concatenated context representations with embeddings as in FVTA network. \textbf{\textit{TGIF Temporal Attention}} \cite{jang2017tgif} is a recently proposed spatial-temporal reasoning network on sequential animated image QA. Since other baseline methods do not use spatial attention, we compare the TGIF network with temporal attention only. TGIF temporal attention uses a simple MLP to compute the attention and only the last hidden state of the question is considered. We compute the attention following \cite{jang2017tgif} and use the same output layer in our method.


\noindent\textbf{Implementation Details}
\label{sec-impl}
In MemexQA dataset, each question is asked to a sequence of photos organized in albums. A photo might have 5 types of textual metadata, including the \textit{album title}, \textit{album descriptions},  \textit{GPS Locations}, \textit{timestamp} and a \textit{title}. We use  $N$ to denote the maximum number of albums, $K$ for the maximum number of photos in an album and $V$ for the maximum words. For album-level textual sequences like album titles and descriptions, the $K$ dimension only has one item and others are zero-padded. We also use zeros to pad those positions with no word/image.
We encode GPS locations using words.
The photos and their corresponding metadata form the visual-text sequences. All questions, textual context and answers are tokenized using the Stanford word tokenizer. We use pre-trained GloVe word embeddings \cite{pennington2014glove}, which is fixed during training.
For image/video embedding, we extract fixed-size features using the pre-trained CNN model, Inception-ResNet \cite{szegedy2017inception}, by concatenating the pool5 layer and classification layer's output before softmax. We then use a linear transformation to compress the image feature into 100 dimensional.
Then a bi-directional LSTM is used for each modality to obtain contextual representations. Given a hidden state size of $d$,  which is set to 50, we concatenate the output of both directions of the LSTM and get a question matrix $\mathbf{Q} \in \mathbb{R}^{2d \times M}$
and context tensor $\mathbf{H} \in \mathbb{R}^{2d \times V \times K \times N \times 6}$ for all media documents. 
We reshape the context tensor into $\mathbf{H} \in \mathbb{R}^{2d \times T \times 6}$. 
To select the best hyperparmeters, we randomly select 20\% of the official training set as the validation set.
We use the AdaDelta \cite{zeiler2012adadelta} optimizer and an initial learning rate of 0.5 to train for 200 epochs with a dropout rate of 0.3.

\subsubsection{Comparison to the state-of-the-art} \label{FVT_comp_exp}

Table~\ref{exp-mem} compares the accuracy on the MemexQA. As we see, the proposed method consistently outperforms the baseline methods and achieves the state-of-the-art accuracy on this dataset. The first 3 methods in the table show the performance of embedding methods without any attentions. Although embedding methods are relatively simple to implement, their performance is much lower than the proposed FVTA model. The experiment results advocate the attention model among images and image sequences. Compare to previous attention models, our FVTA network significantly outperforms other methods, which proves the efficacy of the proposed method.


\begin{table}[ht]
\centering
\begin{tabular}{l||c|c|c}
\hline
               & HIT@1 & HIT@3 & mAP \\ \hline \hline
Soft Attention    & 1.16\%         & 12.60\% & 0.168$\pm$0.002                      \\
MCB    & 11.98\%         & 30.54\% & 0.269$\pm$0.005                      \\
TGIF Temporal    & 13.28\%         & 32.83\% & 0.289$\pm$0.005                      \\
FVTA & \textbf{15.48}\%          & \textbf{35.66}\%    & \textbf{0.312$\pm$0.005}                  \\
\hline
\end{tabular}
\vspace{1mm}
\caption{The quality comparison of the learned FVTA and classic attention. We compare the image of the highest activation in a leaned attention to the ground truth evidence photos which human used to answer the question. HIT@1 means the rate of the top attended images being found in the ground truth evidence photos. AP is computed on the photo ranked by their attention activation.}
\label{att-acc}
\vspace{-3mm}
\end{table}

\begin{figure*}[!t]
	\centering
		\includegraphics[width=0.99\textwidth]{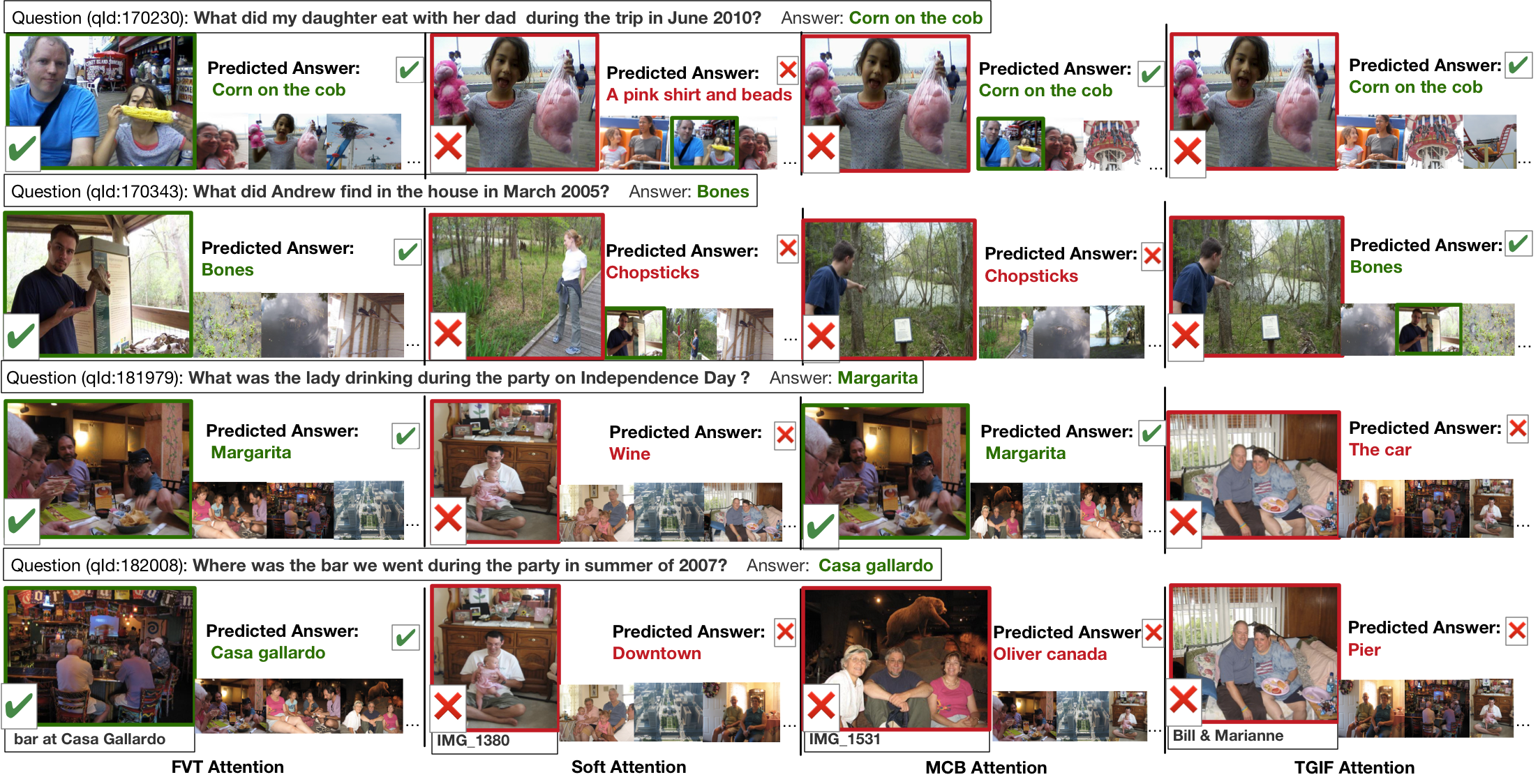}
	\caption{ 
	Qualitative comparison of FVTA model and other attention models on the MemexQA dataset. For each question, we show the answer and the images of the highest attention weights. Images are ranked from left to right based on the attention weights. The correct images and answers have green border whereas the incorrect ones are surrounded by the red border. 
	}
	\label{FVT_comp_vis}
\end{figure*}

The MemexQA dataset provides ground truth evidence photos for every question. We can compare the correlation between the photos of the highest attention weights and the ground truth photos to correctly answer a question. An ideal VQA model should not only enjoy a high accuracy in answering a question (Table~\ref{exp-mem}) but also can find images that are highly correlated to the ground-truth evidence photos. Table~\ref{att-acc} lists the accuracy to examine whether a model puts focus on the correct photos. FVTA outperforms other attention models on finding the relevant photos for the question. The results show that the proposed attention can capture salient information for answering the question. For qualitative comparison, we select some representative questions and show both the answer and the retrieved top images based on the attention weights in Fig.~\ref{FVT_comp_vis}. 
As shown in the first example, the system has to find the correct photo and visually identify the object to answer the question "what did the daughter eat while her dad was watching during the trip in June 2010?". FVTA attention puts a high weight on the correct photo of the girl eating a corn, which leads to correctly answering the question. Whereas for soft attention, the one-dimensional attention network outputs the wrong image and gets the wrong answer.
This example shows the advantage of FVTA modeling the correlation at every time step, across visual-text sequences over the traditional dimensional attention. 

\subsubsection{Ablation Study}
Table \ref{exp-mem-abla} shows the performance of FVTA mechanism and its ablations on the MemexQA dataset. 
To evaluate the FVTA attention mechanism, we first replace our kernel tensor with simple cosine similarity function. 
Results show that standard cosine similarity is inferior to our similarity function. 
For ablating intra-sequence dependency, we use the representations from the last timestep of each context document. For ablating cross sequence interaction, we average all attended context representation from different modalities to get the final context vector.
Both aspects of correlation of the FVTA attention tensor contribute towards the model's performance, while intra-sequence dependency shows more importance in this experiment. We compare the effectiveness of context-aware question attention by removing the question attention and use the last timestep of the LSTM output from the question as the question representation. It shows the question attention provides slight improvement. Finally, we train FVTA without photos to see the contribution of visual information. The result is quite good but it is perhaps not surprising due to the language bias in the questions and answers of the dataset, which is not uncommon in VQA dataset~\cite{antol2015vqa} and in Visual7W~\cite{zhu2016visual7w}. This also leaves significant rooms of improvement with visual information.

\begin{table}[ht]
\centering
\begin{tabular}{l||c|c}
\hline
Ablations      & Accuracy & $\Delta$ \\ \hline \hline
FVTA w/ Cosine Similarity  & 0.619 & -4.9\% \\
FVTA w/o Intra-seq  &0.569& -10.0\% \\
FVTA w/o Cross-seq  & 0.604& -6.5\%\\
FVTA w/o Question Attention 		&0.629 &-4.0\%\\ 
FVTA w/o Photos &	0.577& -9.1\%\\
\hline
\end{tabular}
\vspace{1mm}
\caption{Ablation studies of the proposed FVTA method on the MemexQA dataset. 
The last column shows the performance drop.}
\label{exp-mem-abla}
\end{table}

\begin{figure*}[!t]
	\centering
		\includegraphics[width=0.99\textwidth]{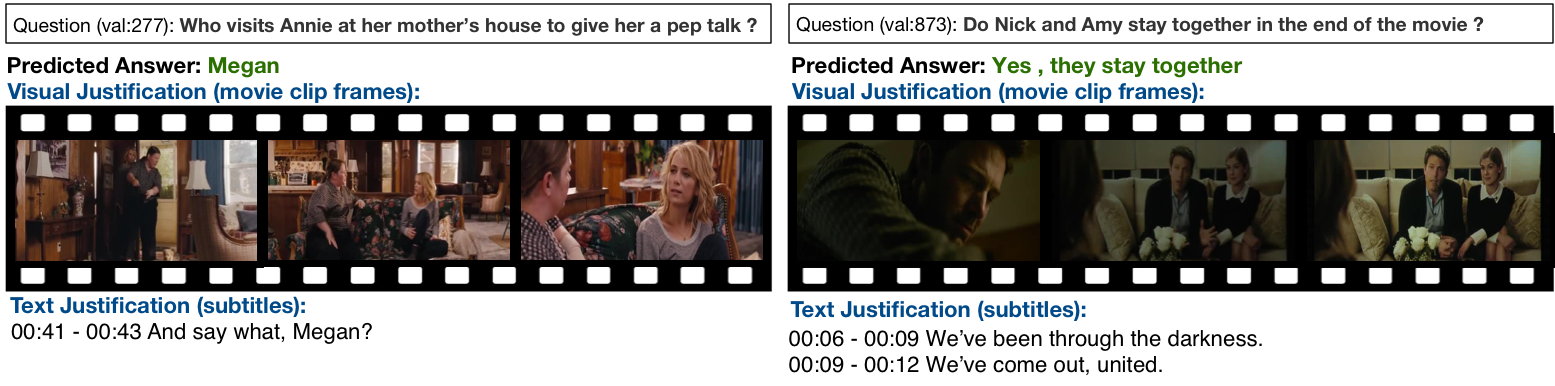}
	\caption{ 
	Qualitative analysis of FVTA on the MovieQA dataset. It shows the visual justification (movie clip frames) and text justification (subtitles) based on the top attention activation. Both justifications provide supporting evidence for the system to get the correct answer.
	}
	\label{FVT_comp_movieqa_vis}
\end{figure*}

\subsection{MovieQA}
\noindent\textbf{Dataset}
The MovieQA dataset consists of 140 movies and 6,462 multiple choice QA pair. Each QA pair contains five answer choices with only one correct answer. Systems are required to answer the questions given a number of movie clips from the same movie and the corresponding subtitles. 
More details of the dataset can be viewed in~\cite{tapaswi2016movieqa}.

\noindent\textbf{Implementation Details}
\label{sec-impl2}
In the MovieQA dataset, each QA is given a set of $N$ movie clips of the same movie, and each clip comes with subtitles. We implement FVTA network for MovieQA task with modality number of 2 (video \& text). 
We set the maximum number of movie clips per question to $N=20$, the maximum number of frames to consider to $F=10$, the maximum number of subtitle sentences in a clip to $K=100$ and the maximum words to $V=10$. Visual and text sequences are encoded the same way as in the MemexQA \cite{jiang2017memexqa} experiment.  
We use the AdaDelta \cite{zeiler2012adadelta} optimizer with a minibatch of 16 and an initial learning rate of 0.5 to trained for 300 epochs. A dropout rate is set at 0.2 during training. 
The official training/validation/test split is used in our experiments.

\begin{table}[]
\centering
\begin{tabular}{l||c|c}
\hline
Method            & Val & Test \\
\hline \hline
SSCB \cite{tapaswi2016movieqa}          & 0.219 & -\\
MemN2N \cite{tapaswi2016movieqa}        & 0.342 & -\\
DEMN \cite{kim2017deepstory} & - & 0.300\\
Soft Attention & 0.321 & -\\
MCB \cite{fukui2016multimodal}& 0.362 & -\\
TGIF Temporal \cite{jang2017tgif} & 0.371 & -\\
RWMN \cite{na2017read} & 0.387 & 0.363 \\
FVTA        & \textbf{0.410} & \textbf{0.373} \\
 \hline
\end{tabular}
\vspace{1mm}
\caption{Accuracy comparison on the test and the validation set of the MovieQA dataset. The test set performance can only be evaluated on the MovieQA server, and thus not all the studies provide the accuracy on Test set.
}
\label{exp-movieqa}
\end{table}

\noindent\textbf{Experimental Results}
We compare FVTA with recent results on MovieQA dataset, including End-to-End Memory Network (MemN2N) \cite{MovieQA}, Deep Embedded Memory Network (DEMN) \cite{kim2017deepstory}, and Read-Write Memory Network (RWMN) \cite{na2017read}. 
Table~\ref{exp-movieqa} shows the detailed comparison of MovieQA results using both videos and subtitles. FVTA model outperforms all baseline methods and achieves 
comparable performance to the state-of-the-art result \footnote{The best test accuracy on the leaderboard by the time of paper submission (Nov. 2017) is 0.39 (Layered Memory Networks). It is not included in the table as there is no publication to cite.} on the MovieQA test server. Notably, RWMN \cite{na2017read} is a very recent work that uses memory net to cache sequential input, with a high capacity and flexibility due to the read and write networks. Our accuracy is 0.410 (vs 0.387 by RWMN) on the validation set and 0.373 (vs 0.363) on the test set. Benefiting from such modeling ability,
FVTA consistently outperforms the classical attention models including soft attention, MCB \cite{fukui2016multimodal} and TGIF \cite{jang2017tgif}. The result demonstrates the consistent advantages of FVTA over other attention models in question-answering for multiple sequence data.

Fig.~\ref{FVT_comp_movieqa_vis} illustrates the output of our FVTA model. FVTA can not only predict the correct answer, but also identify the most relevant subtitle description as well as the movie clip frames. As shown in Fig.~\ref{FVT_comp_movieqa_vis}, FVTA can provide fine-grained level justifications such as the most informative movie frames or subtitle sentences, whereas most of existing methods cannot find fine-grained justifications from the attention computed at the movie clip level. We believe the results show the benefits and potentials of FVTA model.

\section{Conclusions and future work}
In this paper, we introduced a novel neural network model called Focal Visual-Text Attention network for answering questions over visual-text sequences. FVTA employed a hierarchical process to dynamically determine which modality and snippets to focus on in the sequential data to answer the question, and hence can not only predict the correct answers but also find the correct supporting justifications to help users verify the system's results.
The comprehensive experimental results demonstrated that FVTA achieves comparable or even better than state-of-the-art results on two major question answering benchmarks of sequential visual-text data. Our future work includes extending FVTA to large scale long visual-text sequences and removing the use of answer choice embeddings as the input.



\noindent\textbf{Acknowledgements} We would like to thank anonymous reviewers for useful comments and Google Cloud for providing GCP research credits.

{\small
\bibliographystyle{ieee}
\bibliography{egbib}
}

\end{document}